%% file: mersch2021iros.tex
\documentclass[letterpaper, 10 pt, conference]{ieeeconf}
\IEEEoverridecommandlockouts    %
\input{stachnisslab-latex}

\input{stachnisslab-math}

\usepackage{multirow}

\usepackage{xcolor}
\usepackage{transparent}
\usepackage{graphicx}
\graphicspath{{pics/}}

\title{\LARGE \bf Maneuver-based Trajectory Prediction for Self-driving Cars \\Using Spatio-temporal Convolutional Networks\\}
\author{Benedikt Mersch$^{1,*}$ \qquad Thomas H\"ollen$^{1,*}$ \qquad Kun Zhao$^2$ \qquad Cyrill Stachniss$^1$ \qquad Ribana Roscher$^1$
\thanks{$^*$ These authors contributed equally.}
\thanks{$^1$ University of Bonn, Germany}
\thanks{$^2$ Aptiv Services Deutschland GmbH, Wuppertal, Germany}
\thanks{This work has partially been funded by the Deutsche Forschungsgemeinschaft (DFG, German Research Foundation) under Germany's Excellence Strategy, EXC-2070 - 390732324 - PhenoRob and by the European Union’s Horizon 2020 research and innovation programme under grant agreement No~101017008~(Harmony).}%
}

\begin{document}
\maketitle
\thispagestyle{empty}
\pagestyle{empty}

\begin{abstract}

The ability to predict the future movements of other vehicles is a subconscious and effortless skill for humans and key to safe autonomous driving. Therefore, trajectory prediction for autonomous cars has gained a lot of attention in recent years. It is, however, still a hard task to achieve human-level performance. Interdependencies between vehicle behaviors and the multimodal nature of future intentions in a dynamic and complex driving environment render trajectory prediction a challenging problem.
In this work, we propose a new, data-driven approach for predicting the motion of vehicles in a road environment. The model allows for inferring future intentions from the past interaction among vehicles in highway driving scenarios. Using our neighborhood-based data representation, the proposed system jointly exploits correlations in the spatial and temporal domain using convolutional neural networks. Our system considers multiple possible maneuver intentions and their corresponding motion and predicts the trajectory for five seconds into the future. We implemented our approach and evaluated it on two highway datasets taken in different countries and are able to achieve a competitive prediction performance.

\end{abstract}

\section{Introduction}
\label{sec:intro}
Being able to predict the future movement of other traffic participants is important when driving on roads. Experienced human drivers subconsciously anticipate possible maneuvers of other road users and react accordingly in advance. Especially with varying numbers and types of traffic participants in dynamic and complex road environments, predicting the intentions of other drivers is a challenging task and humans refine this skill over time while typically outperforming technical systems. 

Intelligent vehicles require a reliable perception of the driving environment over time and space to make medium- or even long-term predictions. This is not only the case for autonomous driving, but also for advanced driver assistance systems. For example, adaptive cruise control or traffic jam assistants need to predict the behaviors of other road users in advance to help the system react on time. The longer the prediction horizon, the more time the system has to provide a safe and at the same time convenient user experience. 

\begin{figure}[t]
  	\centering
  	\vspace{0.2cm}
  	\fontsize{9}{9}\selectfont
	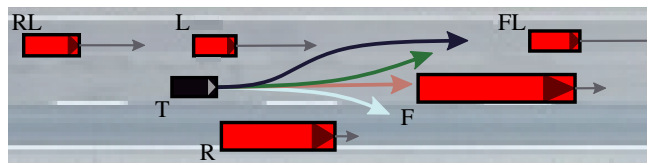
  	\caption{The goal is to predict the future maneuver of the black target vehicle (T). The different future motions are estimated depending on the neighbors' states like position, velocity, and acceleration. Colors indicate possible maneuvers with corresponding trajectories. Gray arrows denote velocity vectors, their lengths indicate speed.}
	\label{fig:motivation}
\end{figure}

In contrast to pedestrians and bicycles, cars are more restricted in their motion due to larger inertia, traffic rules, and road geometries. This makes vehicle movements better predictable, especially in structured driving scenarios like highway driving. Despite these amendable properties, challenges arise due to the dynamic interactions among vehicles. Particularly lane change maneuvers require special attention for surrounding objects. Given a so-called target vehicle, for which one wants to predict the future trajectory, neighboring vehicles influence but also restrict the possible maneuvers, such as accelerating or changing a lane. The situation depicted in~\figref{fig:motivation} illustrates such a scenario. In case the black car (T) is faster than the truck in front (F), it will need to slow down or make a lane change. Possible scenarios are shown as colorized trajectories. Their likelihood strongly depends on the positions, velocities, and accelerations of the neighboring cars F, L, R, RL, and FL. The example illustrates the inherent interdependencies among vehicle behaviors for a time horizon of several seconds. This paper proposes a method to tackle these challenges by paying attention to the other road users and their dynamics and by explicitly estimating the driver's maneuver intention in advance.

The main contributions of this paper towards vehicle motion prediction are two-fold. First, we present a novel semantic neighborhood representation of the scene around a target vehicle for joint aggregation of higher-level features in prediction tasks. The memory-efficient and dense 3D tensor encodes the time, neighbor positions, and past vehicle states as the dynamic context. Second, we propose the use of two 2D convolutional neural networks (CNNs) for joint spatio-temporal feature extraction from the proposed input representation. Our approach explicitly uses convolutions across time and the space of neighboring vehicles. To this end, we classify the future lane change intention of a target vehicle with respect to a lateral motion and then predict a trajectory based on the classified lane change intention. Using a spatio-temporal CNN for sequence prediction leads to a simpler and more compact architecture compared to recurrent network approaches resulting in fewer parameters to train but competitive prediction performance.

This yields an approach that is able to
(i) successfully classify lane change maneuvers and predict a corresponding trajectory for real-world scenarios by
(ii) performing joint spatio-temporal feature aggregation with 2D CNNs,
(iii) and outperform state-of-the-art methods.
These three main claims are backed up by the paper, our experimental evaluation, and an ablation study.

\section{Related Work}
\label{sec:related}

Brown~\etal~\cite{brown2020arxiv} break down the task of estimating and predicting human driver behavior into the tasks of state estimation, intention estimation, trait estimation, and motion prediction. In this work, we focus on predicting the future motion of a vehicle based on the estimated driver intention.

Inferring the future motion of a vehicle from past data has been approached from different perspectives. Closed-loop approaches roll out a control policy in a forward simulation for all target vehicles up to the prediction time, which results in interaction-aware trajectories~\cite{schulz2018iros}. Interaction-awareness between different traffic participants is deepened with game-theoretic approaches~\cite{sun2019iv}, which condition an agent's future motion on the predicted motion of others. Since this dependency increases the computational complexity and can become intractable, other approaches model the future behaviors of vehicles to be independent of each other. Among these, physics-based models solely use kinematic and dynamic properties and apply filter- or sampling-based methods~\cite{lefevre2014robo}. More advanced independent prediction models are data-based approaches, which are further addressed in this work.

Mozaffari~\etal~\cite{mozaffari2020arxiv} differentiate data-based prediction approaches based on input representation, output type, and prediction method. If only the target's trajectory history is used for prediction, interdependencies between surrounding vehicles can not be considered~\cite{xin2018itsc,park2018iv}. In contrast, adding information about neighboring agents~\cite{deo2018iv,dai2019acce}, stacking different sources of spatial information in Bird's Eye Views (BEV)~\cite{cui2019icra,hoermann2018icra} or using raw sensor data of the target's surroundings makes interaction-awareness between past states possible~\cite{casas2018corl,li2020iros}. Deo and Trivedi~\cite{deo2018cvprws} use a combination of these approaches and call it a social tensor. This tensor consists of a BEV image of the scene and is augmented by pre-processed temporal features. In this paper, we extend this idea by defining a spatio-temporal representation in advance and then jointly aggregating spatial and temporal features for prediction instead of processing them separately~\cite{deo2018cvprws}.

A common goal of motion prediction is the estimation of maneuver intentions. These methods provide a high-level understanding of future behavior, usually defined for specific scenarios like intersections~\cite{zyner2018tits} or highway lane changes~\cite{ding2019icra}. To predict low-level future motion, an occupancy grid map containing the probabilities of occupancy at future time steps can be used~\cite{hoermann2018icra}. This output representation also allows for predicting multiple future modes but lacks accuracy for large grid cells resulting in less consistent trajectories. Another possibility is to predict trajectories directly, either in a uni- or multimodal fashion. Unimodal predictions come with a lower computational cost but tend to converge to a mean of different behavior modes~\cite{dai2019acce,luo2018cvpr}. Maneuver-based predictions fix this problem by outputting multiple motion hypotheses incorporating different maneuvers. These models can be formulated in a probabilistic framework, meaning that they model or sample from a multimodal distribution conditioned on the input data~\cite{zhao2019cvpr,gupta2018cvpr,tang2019neurips}. Other approaches estimate intention modes in advance and use them to predict a trajectory~\cite{song2020eccv,xin2018itsc} based on the maneuver. Casas~\etal~\cite{casas2018corl} perform a multi-class classification with eight intention classes using CNNs on a voxelized LiDAR scan and a dynamic map containing road structures and traffic lights. The generated intention scores are then further processed to condition the trajectory estimation. In contrast to these approaches, we define an intention space with three lateral motions and classify a maneuver at each step. Combining these motions at each step results in a larger variety of maneuvers during the prediction.

Different methods have evolved for processing the spatial and temporal dependencies to predict future behavior. Recurrent neural networks (RNNs) can predict time series by maintaining a hidden state while processing temporal information. However, they can be difficult to train for larger time series due to vanishing or exploding gradients caused by their recurrent structure. Therefore, more advanced recurrent models like gated recurrent units (GRUs)~\cite{chung2014nipsws} or long short-term memory networks (LSTMs)~\cite{hochreiter1997neuralcomputation} have been proposed and applied for trajectory prediction~\cite{alahi2016cvpr,gupta2018cvpr,kim2017itsc,amirian2019cvprws}. Since only extracting temporal information does not account for dependencies between traffic participants, convolutional neural networks (CNNs) have been widely implemented for spatial information aggregation~\cite{hoermann2018icra,casas2018corl,hong2019cvpr,cui2019icra}. Other methods are fully connected~\cite{hu2018iv} or graph neural networks~\cite{diehl2019iv}. Generative adversarial methods (GANs) proposed by Goodfellow~\etal~\cite{goodfellow2014nips} have also been considered for trajectory prediction~\cite{kuefler2017iv,zhao2019cvpr}. The already mentioned approach by Deo and Trivedi~\cite{deo2018cvprws} uses a combination of LSTMs for temporal feature extraction for each vehicle trajectory and CNNs for processing the resulting grid representation. The final trajectory is generated by an LSTM decoder for each maneuver. Other combinations of the presented approaches can be found for trajectory prediction~\cite{sadeghian2019cvpr,zhao2019cvpr,bansal2019rss}. The main difference to our approach is that we solely use a convolutional architecture for jointly encoding temporal and spatial information as well as decoding the resulting features for prediction without the use of recurrent structures.

The use of CNNs instead of RNNs for sequence modeling has been proposed by Bai~\etal~\cite{bai2018arxiv}. They argue that such temporal convolutional networks (TCN) are capable of outperforming recurrent models for specific tasks while being simpler and easier to train. In addition to that, TCNs can process the data in parallel, which makes them faster at inference. Nikhil and Morris~\cite{nikhil2018eccvws} propose to use CNNs for pedestrian trajectory prediction and show superior performance and speed compared to LSTM-based models. However, neighboring agents are not considered in their method since they only use 1D convolutions along the time dimension. The approach developed in this work is inspired by temporal convolutions and provides, in contrast to Nikhil and Morris~\cite{nikhil2018eccvws}, a joint aggregation of spatial and temporal features from the proposed novel tensor input representation.

\section{Our Approach}
\label{sec:main}

The key idea of our approach is to first classify the lane change maneuver of a target vehicle for each step into the future and then predict the corresponding trajectory with the classification result as an additional input. 

At the current time step $t{=}0$, the goal is to predict the sequence of $P$ future positions $\{(x_t,y_t)\}_{t=1}^P$ from $H$ past states $\mathcal{S}{=}\{s_{-t}\}_{t=0}^{H-1}$. Each past state $s_{-t}$ consists of multiple channels like position, acceleration, and velocity of the target vehicle and its neighbors, which are transformed into a 3D tensor. This representation is addressed in~\secref{sec:input}. The lane change maneuvers are pre-defined and cover sequences of straight driving, left, and right lane changes as explained in~\secref{sec:output}. The classification and regression steps rely on the exploitation of past state sequences with spatio-temporal 2D convolutions, which are presented in~\secref{sec:cnn}. Based on the CNN features, the classification module predicts a discrete maneuver (straight driving, left lane change, or right lane change) for each prediction step $1,...,P$ into the future, see~\secref{sec:classification}. The classified sequence is then passed to the regression module as an additional input to predict a trajectory based on the maneuver as outlined in~\secref{sec:regression}. Both modules share the same convolution-based architecture. When training on common, publicly available highway datasets, lane changes are usually underrepresented since the vehicles drive straight most of the time. We found that training a single network with two heads and a joint multi-task loss for classification and regression is hard to tune due to the imbalance of lane change labels. Therefore, we propose to use partitioned models for each task, which makes it easier to train them separately without shared parameters.

\subsection{Neighborhood-based Input Representation}
\label{sec:input}

\begin{figure}[t]
	\centering
	\vspace{0.2cm}
	\hspace*{-0.2cm}
	\fontsize{9}{9}\selectfont
	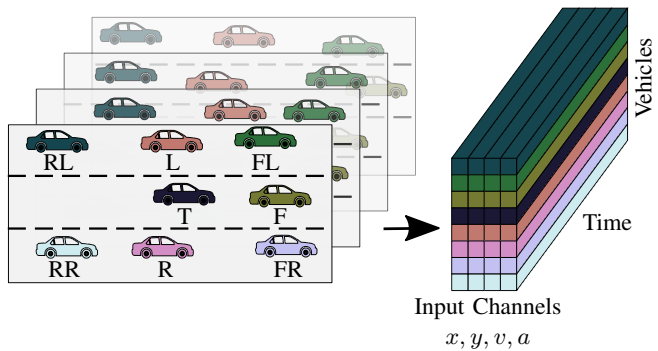
	\caption{\textit{Left}: At each time step, we discretize the neighborhood around the target vehicle into seven neighbor positions. \textit{Right}: For each neighbor, we extract the input channels $x$ and $y$ positions, velocity $v$, and acceleration $a$ for each time step resulting in 2D tensors and concatenate them to a 3D input tensor.}
	\label{fig:input}
	\vspace{-0.2cm}
\end{figure}

We take the direct neighborhood of the target vehicle into account to predict a trajectory from past data. Close neighbors influence the possible actions a vehicle can take. For example, a leading car with a lower velocity requires deceleration or a lane change of the target vehicle, but the corresponding neighboring lane needs to be free for a safe lane change maneuver. Whereas the short-time prediction of for example the next second is mainly restricted by the inertia of the car, we assume that for longer prediction horizons, including the neighborhood leads to better predictions. This claim is backed up by our experiment in~\secref{sec:ablation}. We define a semantic neighborhood consisting of up to seven vehicles around the target inspired by the work of Hu~\etal~\cite{hu2018iv}. As mentioned, the vehicles in front (F) and on the left (L) and right (R) play an important role. The remaining neighborhood is further discretized in front left (FL) and right (FR) as well as rear left (RL) and right (RR) vehicles. In total, this results in a maximum number of eight considered vehicles including the target. We illustrate the semantic neighborhood in~\figref{fig:input}. For each neighbor vehicle, we acquire an image-like 2D tensor containing the different input channels over time and concatenate them to a 3D tensor. In case a neighbor is not present, we set the entries to zero, which is a state that is not naturally present in the input data due to the choice of the reference location. The concatenation order of neighbors can be chosen randomly but needs to be consistent for training and evaluation. We experienced that it is advantageous to keep neighbor positions close to each other in the data representation. Based on our experiments in~\secref{sec:ablation}, we hypothesize that the convolutional network can learn a notion of absolute position and discriminates between neighbors. This is further discussed in~\secref{sec:cnn}. In contrast to other approaches that encode the whole scene into a grid structure~\cite{deo2018cvprws}, our neighborhood representation is more dense and memory-efficient, since not all vehicles are taken into account.

The proposed multi-channel design makes it easy to add new information channels that can be useful for prediction. We use the information given by the publicly available highway datasets, which are $x$ and $y$ position, velocity, and acceleration. Note that velocity and acceleration can also be extracted from the positional information over time in case they are not directly provided. We normalize the data beforehand to guarantee a consistent input scaling.

We store the neighbors and channels per frame along the third, temporal dimension, see~\figref{fig:input}. In order to compare our results with existing methods, we use three seconds of input data and predict the maneuvers and positions for the following five seconds. In this setup with data sampled at 10\,Hz, this results in an input tensor of dimension $4{\times}8{\times}30$ with 4 channels, 8 neighbors, and 30 frames of history.

\begin{figure*}[t]
	\centering
	\vspace{0.2cm}
	\fontsize{9}{9}\selectfont
	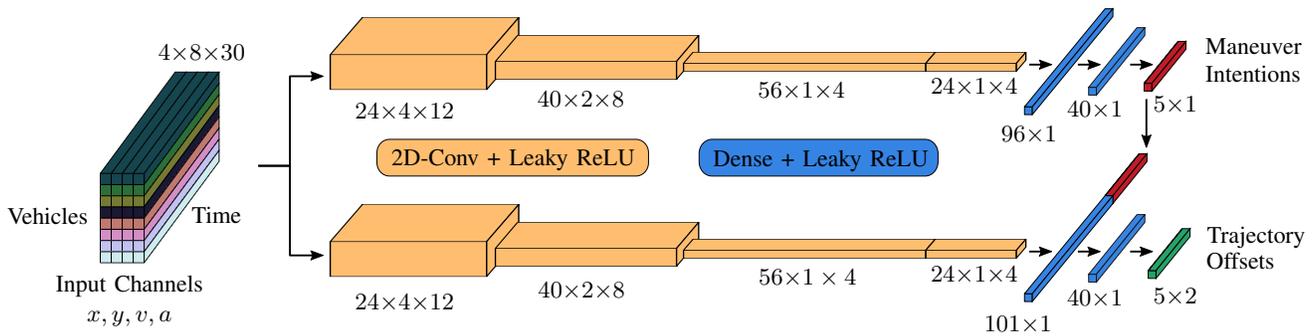
	\caption{Two spatio-temporal convolutional networks aggregate features in parallel from the neighborhood-based 3D input tensor. The tensor from ~\figref{fig:input} contains the $x$ and $y$ positions, velocity $v$ and acceleration $a$ for all eight vehicles at the past time steps. Note that the transposed input tensor is depicted for better visualization. The upper module classifies a maneuver for each prediction step and passes the result to the lower regression module. Finally, the offsets between the future trajectory and the initial position are predicted. For each layer we provide the output dimensions.}
	\label{fig:architecture}
	\vspace{-0.2cm}
\end{figure*}

\subsection{Output Parameterization}
\label{sec:output}
The number of predicted future positions $P$ depends on the output size of the last dense layer in the classification and regression heads in~\figref{fig:architecture} and is only limited by the resolution of the dataset used for training. We choose to predict and evaluate the future for the next 5\,s represented via 5 positions being 1\,s apart in time. This sampling rate is commonly used for evaluating the performance of trajectory prediction methods~\cite{deo2018cvprws},~\cite{zhao2019cvpr},~\cite{song2020eccv}, and therefore, we use it here. Note that the input data is still processed at 10\,Hz, which is unaffected by the output resolution. Furthermore, a higher output-rate can be easily added. A ground truth maneuver output with 3 classes (straight, left lane change, right lane change) is of size $5{\times}3$ and is computed from the lane IDs. The lane ID is provided by the datasets and not part of the input data. If the lane ID of a vehicle changes between subsequent frames, we assume a lane change with a duration of 2\,s before and after the corresponding frames and infer the maneuver direction from the lane IDs.

The regression layer outputs trajectory offsets that describe the difference between the predicted position and the initial position at $t{=}0$ of the target vehicle. This improves the training since the identity mapping from input to output is achieved by predicting an offset of zero. At inference time, the model's output is de-normalized and added to the target's current position.

\subsection{Spatio-temporal Convolutional Neural Network}
\label{sec:cnn}
Given the data representation proposed in~\secref{sec:input}, we use four convolutional layers for feature aggregation. As carried out by Bai~\etal~\cite{bai2018arxiv}, we apply convolutions along the time dimension to detect local correlations between states in the history sequence. In addition to that, we extend the temporal convolution along the spatial neighbor dimension resulting in 2D convolutions. The hierarchical stacking of layers ensures that the low-level features from the input data can be combined in the subsequent layers for higher-level representations. We conclude that higher layers learn to interpret the convolution along the neighborhood dimension and therefore account for the absolute position of the neighbor. This assumption is backed up by Kayhan~\etal~\cite{kayhan2020cvpr}, who claim that CNNs can encode absolute spatial locations from boundary effects. Since we use zero padding and our spatial dimension is only eight, we can expect to make use of the boundary effects during training.

It is worth pointing out that the learning process of the temporal and spatial relations among all vehicles happens jointly. This makes the prediction fast since, in contrast to recurrent models, no previous hidden states need to be computed and taken into account. Analogously to vision-based 2D convolutions with multiple image channels, the filter depth is matched with the number of feature channels.

Stacking multiple convolutional layers increases the receptive field of the CNN. For temporal convolutional networks, the receptive field determines the number of past time steps that influence a single output. Especially for longer input sequences, this is an important factor for prediction performance. Besides deepening the architecture by adding layers, it is also possible to increase the size of the convolution kernels. However, since all techniques increase the number of training parameters and therefore the model complexity, we use dilated convolutions. A dilated convolution expands the receptive field without reducing the resolution or coverage as pointed out by Yu and Koltun~\cite{yu2016iclr}. We implement dilations along the temporal dimension by adding zeros between kernel entries depending on the dilation rate. We provide an experiment on the improvement by using dilated convolutions in~\secref{sec:ablation}.

The complete network architecture is illustrated in~\figref{fig:architecture}. Each convolutional layer uses a leaky rectified linear unit (ReLU) activation function~\cite{maas2013icml}. At the first layer, the input tensor of size $4{\times}8{\times}30$ is convolved with 24 filters with a kernel size of $5{\times}10$ and dilation of one. Each output in the resulting feature map of size $24{\times}4{\times}12$ has a receptive field of $4{\times}5{\times}19$. Furthermore, a dilated convolution with 40 filters of size $3{\times}3$ is carried out resulting in a size of $40{\times}2{\times}8$. The third layer applies 56 filters of size $2{\times}3$ with dilation, which leads to a map with $56{\times}1{\times}4$ features. Finally, the last convolutional layer reduces the channel dimension with 24 $1{\times}1$ kernels to 24 feature channels and a final receptive field of $4{\times}8{\times}27$.

\subsection{Classification}
\label{sec:classification}
When only using a regression module to predict the trajectory, the model needs to capture multiple possible maneuvers as motivated in~\figref{fig:motivation}. Mozaffari~\etal~\cite{mozaffari2020arxiv} point out that this can lead to predictions that are averaged over all possible modes. We decide to first classify the future maneuver of the target vehicle and to predict a trajectory that depends on the estimated maneuver.

To classify the future maneuver of the vehicle with our spatio-temporal CNN defined in~\secref{sec:cnn}, we flatten the output of the last convolutional layer with a size of $24{\cdot}4{=}96$ and pass it to a fully connected layer with a leaky ReLU activation function and a hidden dimension of $40$. A final fully connected layer processes the hidden feature vector of length $40$. The output layer predicts the class logits, which can be normalized to a probability for each of the three maneuvers "straight", "left", and "right" at each prediction step resulting in an output vector of size $5{\times}3$. Selecting the indices of the maneuvers with the highest predicted probability results in an estimated maneuver intention vector of size $5{\times}1$ containing values between 0 and 2. Based on the three possible maneuvers at each of the \mbox{$P{=}5$} prediction steps, there are in total $3^5{=}243$ maneuver sequences the classification network can model. During training, we optimize the network parameters by minimizing the sum of negative log-likelihoods at each prediction step reading

\begin{equation}
	\mathcal{L}_\text{class}=\sum_{t=1}^{P} -\log{p(c_t)},
\end{equation}
\noindent 
with $p(c_t)$ denoting the predicted probability for the ground truth class $c_t$ at time step $t$. We analyze the improvements on the prediction performance achieved by our classification module in~\secref{sec:ablation}.

\subsection{Regression}
\label{sec:regression}
To make maneuver-based predictions, we feed the classified maneuver sequence directly to the regression module. As discussed above, there are $243$ possible combinations of maneuver predictions for a given input tensor. We concatenate the resulting vector of size $5{\times}1$ containing the indices of the five most likely maneuvers at each time step with the flattened CNN output vector resulting in 101 features. We pass this vector to a fully connected layer with the same architecture as in the classification module. The corresponding hidden vector of size $40$ is the input for the final output layer with a linear activation function, which then predicts a vector of size $5{\times}2$ containing the $x$ and $y$ positions at each of the $5$ prediction steps. During training, we use the ground truth maneuvers as input for the regression module, whereas during evaluation, we use the maneuvers predicted by the classification module. The training aims at minimizing the batch-wise root mean squared error (RMSE) reading
\begin{equation}
	\mathcal{L}_\text{reg}=\sqrt{\frac{1}{P}\sum_{t=1}^{P}\norm{\hat{\v{x}}_t-\v{x}_t}^2},
\end{equation}
\noindent
with predicted locations $\hat{\v{x}}_t$ and ground truth locations $\v{x}_t$ for each time step $t$, where each location consists of the $x$ and $y$ positions.

With the proposed architecture, the temporal and spatial information of the highway scene around a target vehicle is jointly encoded by applying 2D convolutions. As pointed out by Bai~\etal~\cite{bai2018arxiv}, the temporal convolution avoids a long backpropagation path resulting in a more compact model, which is easier to train compared to RNNs. Also, the prediction can be carried out in parallel due to the temporal convolutional structure and is therefore fast. The total amount of trainable parameters for the design depicted in~\figref{fig:architecture} is 65,721. For comparison, the convolutional social pooling approach~\cite{deo2018cvprws} has 194,954 and the multiple futures prediction approach~\cite{tang2019neurips} has 1,073,644 trainable parameters.

\begin{figure*}[t]
	\centering
	\vspace{0.2cm}
	\fontsize{9}{9}\selectfont
	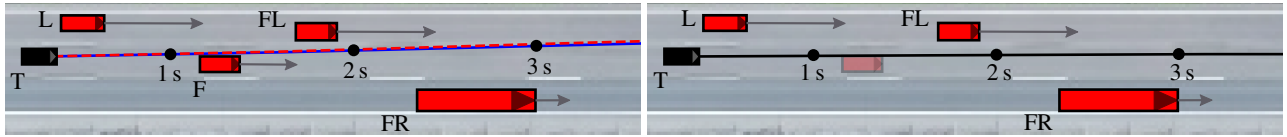
	\caption{Two scenarios with predictions for the black target vehicle (T). \textit{Left}: The original scene with a dashed red line as ground truth. The solid line represents the first seconds of the predicted future trajectory. The velocity of neighboring vehicles is depicted by the length of the gray velocity vectors. Our approach successfully classifies a left lane change (blue) since the car in front (F) is slower. \textit{Right}: In case we modify the neighborhood by removing the car in front, the prediction changes to a straight driving maneuver (black).}
	\label{fig:qualitative_0}
\end{figure*}

\begin{figure*}[t]
	\centering
	\fontsize{9}{9}\selectfont
	\vspace{-0.2cm}
	\hspace*{0.2cm}
	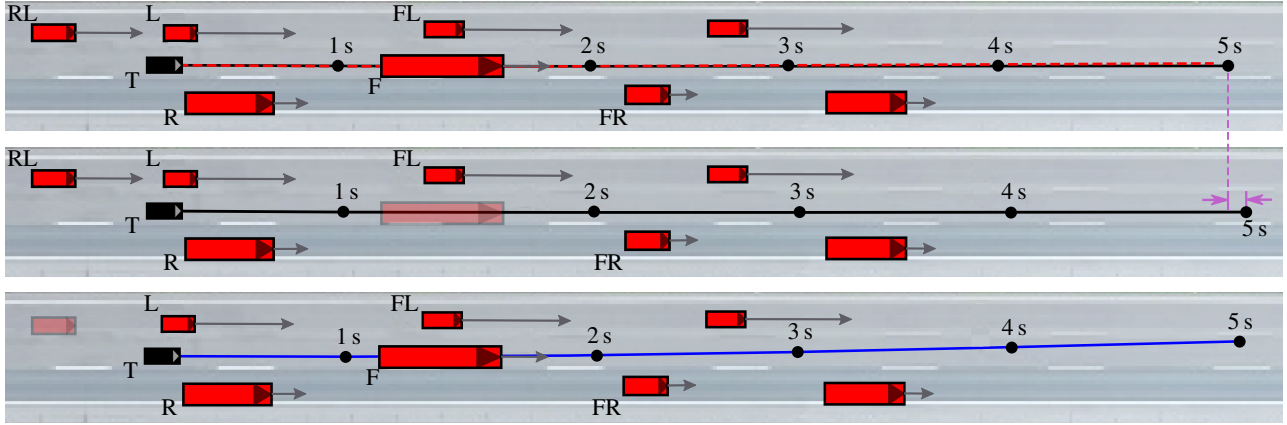
	\caption{Three scenarios with modified environments. \textit{Top}: Original highD scene, the target can not change the lane and needs to decelerate. \textit{Middle}: Removed leading vehicle, the target does not need to decelerate resulting in a longer trajectory. \textit{Bottom}: Removed left rear neighbor, a left lane change (blue) is now possible and therefore predicted. Note that the velocity of neighboring vehicles is depicted by the length of the gray velocity vectors.}
	\label{fig:qualitative_1}
	\vspace{-0.2cm}
\end{figure*}

\section{Experimental Evaluation}
\label{sec:exp}

The main focus of this work is to predict the lane change maneuver and the corresponding future trajectory of a vehicle based on past information about states including the neighborhood.
We present our experiments to show the capabilities of our method and to support our key claims made about our work, which are:
(i) Successfully classifying lane change maneuvers and predicting a corresponding trajectory for real-world scenarios by
(ii) performing joint spatio-temporal feature aggregation with 2D CNNs,
(iii) and outperforming state-of-the-art methods.

\subsection{Implementation Details}
We train the classification and regression modules with the Adam optimizer~\cite{kingma2015iclr} and a learning rate of $7{\cdot}10^{-5}$. Additionally, we augment the training set by overlapping the input data by 20 frames resulting in 506,180 training examples. This leads to better coverage of different lane change maneuvers in the training data. The total training time for each module is 17.5\,h with 300 epochs on an Intel Xeon W-2133 CPU and an Nvidia Quadro RTX 5000 GPU. It takes around 0.3\,ms on average to classify the future maneuver sequence and to predict the corresponding trajectory.

\subsection{Experimental Setup}

We consider two different datasets to train and evaluate our proposed approach on real-world scenarios. The highD dataset~\cite{krajewski2018itsc} contains 110,000 vehicle trajectories from 60 highway recordings. The highways are located in Germany and have 2 to 3 lanes. To show that our approach generalizes well on different driving behaviors and highway structures, we also use the NGSIM dataset~\cite{halkias2006ngsim} with 9,206 vehicles driving on two different US highways with 5 to 6 lanes. Both datasets provide access to position, velocity, and acceleration. For each dataset, we use 70\% of all vehicle trajectories for training, 10\% for validation, and 20\% for testing. For a better comparison, we use the same vehicle trajectories for testing as Song~\etal~\cite{song2020eccv}.

\subsection{Qualitative Analysis}
The qualitative experiments evaluate the prediction results on real-world data and support the claim that our modular approach of classification and regression results in reasonable predicted maneuver-based trajectories. We show that the prediction relies on the spatio-temporal encoding of the target vehicle's surroundings by modifying the neighborhood leading to different results. The original scenes are taken from recording 56 of the highD dataset.

In~\figref{fig:qualitative_0}, we consider two possible scenarios for the same highway scene. The original neighborhood is depicted on the left. A slower car (F) is in front of the black target vehicle (T) and forces it to change the lane or decelerate. In this case, the target will let the left center (L) vehicle pass and perform a lane change. The ground truth trajectory is represented by a red dashed line. Our approach successfully classifies a left lane change maneuver indicated by the blue color and predicts a trajectory that matches the ground truth. To demonstrate the influence of the semantic neighborhood on the prediction, we remove the leading car (F) by replacing the corresponding entries in the input data with zeros. Based on the new neighborhood, our model predicts a straight driving maneuver since there is no longer a need to change the lane. 

The second example in~\figref{fig:qualitative_1} demonstrates a more advanced highway scene. The black target vehicle (T) is again facing a slower vehicle in front (F). Additionally, two cars at the rear and center of the left lane (RL and L) hinder the target from changing the lane. Our approach predicts that the car will stay on the lane and slow down. The ground truth is shown in dashed red and confirms our prediction. In the middle part of~\figref{fig:qualitative_1}, we removed the truck in front (F) from the spatio-temporal input representation. One can see that the predicted trajectory is still straight, but now longer compared to the previous case. This can be reasoned by the fact that there is no longer a slower leading car that forces the target vehicle to decelerate. In a third scenario, we keep the truck in front (F) and remove the car on the rear left (RL). The classification module now outputs a lane change maneuver to the left (blue) and the corresponding trajectory moves to the left lane. All three examples show that our approach can predict reasonable maneuver intentions and trajectories from the neighborhood of the target vehicle.

\begin{table*}
	\centering
	\vspace{0.2cm}
	\begin{center}
		\begin{tabular}{c|c|cccccc}
			\toprule
			Dataset & Time
			& S-LSTM~\cite{alahi2016cvpr} & CS-LSTM~\cite{deo2018cvprws} & S-GAN~\cite{gupta2018cvpr} & MATF~\cite{zhao2019cvpr}
			& PiP-noPlan~\cite{song2020eccv} & Our Approach\\
			\midrule
			\multirow{5}{*}{NGSIM}
			& 1\,s      & 0.60  & 0.58  & 0.57  & 0.66  & 0.55 &\textbf{0.53} \\
			& 2\,s      & 1.28  & 1.26  & 1.32  & 1.34  & 1.20 &\textbf{1.17} \\
			& 3\,s      & 2.09  & 2.07  & 2.22  & 2.08 & 2.00 &\textbf{1.93}\\
			& 4\,s      & 3.10  & 3.09  & 3.26  & 2.97 & 3.01 & \textbf{2.88} \\
			& 5\,s      & 4.37  & 4.37  & 4.40  & 4.13 & 4.27 & \textbf{4.05} \\ \midrule
			\multirow{5}{*}{highD}
			& 1\,s      & 0.19  & 0.19  & 0.30  & - & 0.18 & \textbf{0.10} \\
			& 2\,s      & 0.57  & 0.57  & 0.78  & - & 0.53 & \textbf{0.21}\\
			& 3\,s      & 1.18  & 1.16  & 1.46  & - & 1.09 & \textbf{0.41}\\
			& 4\,s      & 2.00  & 1.96  & 2.34  & - & 1.86 & \textbf{0.78}\\
			& 5\,s      & 3.02  & 2.96  & 3.41  & - & 2.81 & \textbf{1.34}\\
			\bottomrule
		\end{tabular}
		\caption{Comparison of the root mean squared error at each prediction step evaluated on the NGSIM~\cite{halkias2006ngsim} and highD~\cite{krajewski2018itsc} datasets. Bold numbers indicate the best result. The baseline results are reported by Song~\etal~\cite{song2020eccv}. For the stochastic models S-GAN and MATF, the best root mean squared error among 3 sampled trajectories is reported. Note that for PiP, we refer to the PiP-noPlan implementation for a fair comparison, since the planning coupled module in PiP partially uses ground truth information.}
		\label{table:comparison}
	\end{center}
\end{table*}

\begin{table*}
	\centering
	\vspace{-0.2cm}
	\begin{center}
		\begin{tabular}{c|c|cc|c|cc|c}
			\toprule
			&Only Input&Without	&Shuffled		&Without	&Without		&Sampled	&Full  \\
			Time& $x$ and $y$  	&Neighborhood	&Neighborhood&Dilation	&Maneuver	&Maneuver 	&Approach \\
			\midrule
			1\,s&0.24&\textbf{0.08}	&0.15	&0.15	&0.14&0.14&0.10 \\
			2\,s&0.64&\textbf{0.18}	&0.25	&0.30	&0.26&0.33&0.21 \\
			3\,s&1.24&\textbf{0.39}	&0.43	&0.52	&0.47&0.62&0.41\\
			4\,s&2.02&0.83			&0.82	&0.90	&0.87&1.04& \textbf{0.78} \\
			5\,s&3.00&1.52			&1.40	&1.45	&1.48&1.60& \textbf{1.34} \\
			\bottomrule
		\end{tabular}
		\caption{Ablation study on the highD dataset~\cite{krajewski2018itsc}.}
		\label{table:ablation}
	\end{center}
	\vspace{-0.3cm}
\end{table*}

\subsection{Quantitative Analysis}
We present a second experiment to support the claim that our approach outperforms existing state-of-the-art methods for trajectory prediction. For a fair comparison, we evaluate our approach on the same test sets for each dataset as done by Song~\etal~\cite{song2020eccv}. We report the root mean squared error (RMSE) for each prediction step, which is a common metric for trajectory prediction. We compare the results with the baseline methods S-LSTM~\cite{alahi2016cvpr}, CS-LSTM~\cite{deo2018cvprws}, S-GAN~\cite{gupta2018cvpr}, MATF~\cite{zhao2019cvpr} and PiP-noPlan~\cite{song2020eccv}. In case of S-GAN and MATF which are stochastic models, we take the best root mean squared error among 3 sampled trajectories. We refer to the PiP-noPlan implementation since the planning coupled module in PiP uses ground truth future trajectory of the target vehicle to predict the future trajectory of neighboring cars which would result in an unfair comparison.

The final results are shown in~\tabref{table:comparison}. In general, our method shows better results for both datasets at the prediction steps. Especially for the highD dataset, the proposed approach outperforms other state-of-the-art methods at larger prediction horizons. All methods show larger RMSE values for the NGSIM dataset, which can be explained by larger noise in the data as pointed out by Krajewski~\etal~\cite{krajewski2018itsc}. This noise leads to smaller performance improvements between the baselines and also explains why our method achieves a smaller margin compared to the evaluation on the highD dataset. We conclude that noise in for example the vehicle speed heavily influences the prediction due to the high average speed on highways.

\subsection{Ablation Study}\label{sec:ablation}
Finally, we conduct an ablation study to evaluate how much each proposed component of our method contributes to the performance reported in~\tabref{table:comparison}. We retrain and evaluate our approach with several modifications on the highD~\cite{krajewski2018itsc} dataset since the data is more accurate as previously described. The results are shown in~\tabref{table:ablation}.

In a first ablation experiment, we retrain our method only using $x$- and $y$-position as input channels. The result shown in~\tabref{table:ablation} indicates that adding velocity and acceleration as inputs leads to a major performance improvement, which justifies the multi-channel design of the proposed 3D input tensor.

If the local neighborhood defined in~\secref{sec:input} is ignored for prediction as done by Nikhil and Morris~\cite{nikhil2018eccvws}, the prediction performance is slightly better for short-term predictions, but worse for larger prediction horizons. This indicates that the behavior of other traffic participants mainly influences the future trajectory at larger horizons and should not be ignored. Furthermore, we retrain our approach with a shuffled neighborhood order and the competitive result for prediction steps larger than 1\,s indicates that the CNN is able to learn the absolute position across the vehicle dimension independent of the ordering. 

We conducted an additional ablation study to investigate the effect of dilated convolutions used to increase the receptive field and to reduce the number of parameters. If no dilated convolutions are used, the size of the dense layers in~\figref{fig:architecture} needs to be increased to account for the resulting larger feature maps at the output stage of the CNN. This results in a model with 90,681 parameters. Our experiment shows that the use of dilated convolutions leads to a better performance while having fewer parameters to train.

The last ablation study investigates the effect of our maneuver-based prediction. First, if no maneuvers are used, the performance degrades. We hypothesize that the regression module alone is not capable of capturing all maneuvers and therefore a prior classification improves distinguishing between these modes. Finally, we show that our classification module can predict different maneuvers based on the previous input. We test this by comparing to a model which randomly samples a maneuver sequence from the training distribution and passes it to the regression module. It can be seen that the sampled maneuvers lead to a worse performance compared to our trained classification module.

\section{Conclusion}
\label{sec:conclusion}

In this paper, we presented a novel approach to classify the future lane change maneuver of a target vehicle and predict the corresponding trajectory. Our model operates on a 3D spatio-temporal input representation encoding the neighborhood around the target. We exploit the local correlations in neighboring state sequences with spatio-temporal 2D convolutions resulting in a simple, memory-efficient architecture with fast and parallelized inference. This allows us to successfully account for different possible driving maneuvers which are used for a more informed prediction. We implemented and evaluated our approach on two datasets taken at different locations and provided comparisons to other existing techniques and supported all claims made in this work. The experiments suggest that our approach can successfully predict maneuver-based trajectories and outperforms state-of-the-art methods.

\section*{Acknowledgments}
We thank J. Behley, X. Chen, and L. Wiesmann for fruitful discussions and Song~\etal~\cite{song2020eccv} for providing their data preprocessing code and corresponding test split.

\bibliographystyle{plain}

\bibliography{glorified,new}

\end{document}

%% file: stachnisslab-latex.tex
\usepackage{graphics}           
\usepackage{times}              
\usepackage{amsmath}            
\usepackage{amssymb}            
\usepackage{graphicx}
\usepackage{algorithm}
\usepackage[noend]{algpseudocode}
\usepackage{booktabs}

\usepackage[font=small]{caption}

\def\secref#1{Sec.~\ref{#1}}
\def\figref#1{Fig.~\ref{#1}}
\def\tabref#1{Tab.~\ref{#1}}
\def\eqref#1{Eq.~(\ref{#1})}

\makeatletter
\usepackage{xspace}
\DeclareRobustCommand\onedot{\futurelet\@let@token\@onedot}
\def\@onedot{\ifx\@let@token.\else.\null\fi\xspace}

\def\etal{{et al}\onedot}
\makeatother

\usepackage{array}
\newcolumntype{L}[1]{>{\raggedright\let\newline\\\arraybackslash\hspace{0pt}}m{#1}}
\newcolumntype{C}[1]{>{\centering\let\newline\\\arraybackslash\hspace{0pt}}m{#1}}
\newcolumntype{R}[1]{>{\raggedleft\let\newline\\\arraybackslash\hspace{0pt}}m{#1}}

%% file: stachnisslab-math.tex
\newcommand{\norm}[1]{\lVert#1\lVert}

\renewcommand{\b}[1]{\mbox{\boldmath$#1$}}

\renewcommand{\v}[1]{{\b #1}}

%% file: pics/motivation.pdf_tex
\begingroup%
  \makeatletter%
  \providecommand\color[2][]{%
    \errmessage{(Inkscape) Color is used for the text in Inkscape, but the package 'color.sty' is not loaded}%
    \renewcommand\color[2][]{}%
  }%
  \providecommand\transparent[1]{%
    \errmessage{(Inkscape) Transparency is used (non-zero) for the text in Inkscape, but the package 'transparent.sty' is not loaded}%
    \renewcommand\transparent[1]{}%
  }%
  \providecommand\rotatebox[2]{#2}%
  \newcommand*\fsize{\dimexpr\f@size pt\relax}%
  \newcommand*\lineheight[1]{\fontsize{\fsize}{#1\fsize}\selectfont}%
  \ifx\svgwidth\undefined%
    \setlength{\unitlength}{240.46585083bp}%
    \ifx\svgscale\undefined%
      \relax%
    \else%
      \setlength{\unitlength}{\unitlength * \real{\svgscale}}%
    \fi%
  \else%
    \setlength{\unitlength}{\svgwidth}%
  \fi%
  \global\let\svgwidth\undefined%
  \global\let\svgscale\undefined%
  \makeatother%
  \begin{picture}(1,0.24333234)%
    \lineheight{1}%
    \setlength\tabcolsep{0pt}%
    \put(0,0){\includegraphics[width=\unitlength,page=1]{motivation.pdf}}%
    \put(-1.63399478,-0.35866119){\color[rgb]{0,0,0}\makebox(0,0)[lt]{\begin{minipage}{0.93385015\unitlength}\raggedright \end{minipage}}}%
    \put(0,0){\includegraphics[width=\unitlength,page=2]{motivation.pdf}}%
    \put(0.61439567,0.05807857){\color[rgb]{0,0,0}\makebox(0,0)[lt]{\lineheight{1.25}\smash{\begin{tabular}[t]{l}F\end{tabular}}}}%
    \put(0.29944209,0.00395411){\color[rgb]{0,0,0}\makebox(0,0)[lt]{\lineheight{1.25}\smash{\begin{tabular}[t]{l}R\end{tabular}}}}%
    \put(0.00452047,0.20784679){\color[rgb]{0,0,0}\makebox(0,0)[lt]{\lineheight{1.25}\smash{\begin{tabular}[t]{l}RL\end{tabular}}}}%
    \put(0,0){\includegraphics[width=\unitlength,page=3]{motivation.pdf}}%
    \put(0.26131331,0.20784679){\color[rgb]{0,0,0}\makebox(0,0)[lt]{\lineheight{1.25}\smash{\begin{tabular}[t]{l}L\end{tabular}}}}%
    \put(0,0){\includegraphics[width=\unitlength,page=4]{motivation.pdf}}%
    \put(0.76701539,0.20784679){\color[rgb]{0,0,0}\makebox(0,0)[lt]{\lineheight{1.25}\smash{\begin{tabular}[t]{l}FL\end{tabular}}}}%
    \put(0.22929787,0.07055435){\color[rgb]{0,0,0}\makebox(0,0)[lt]{\lineheight{1.25}\smash{\begin{tabular}[t]{l}T\end{tabular}}}}%
  \end{picture}%
\endgroup%

%% file: pics/neighborhood.pdf_tex
\begingroup%
  \makeatletter%
  \providecommand\color[2][]{%
    \errmessage{(Inkscape) Color is used for the text in Inkscape, but the package 'color.sty' is not loaded}%
    \renewcommand\color[2][]{}%
  }%
  \providecommand\transparent[1]{%
    \errmessage{(Inkscape) Transparency is used (non-zero) for the text in Inkscape, but the package 'transparent.sty' is not loaded}%
    \renewcommand\transparent[1]{}%
  }%
  \providecommand\rotatebox[2]{#2}%
  \newcommand*\fsize{\dimexpr\f@size pt\relax}%
  \newcommand*\lineheight[1]{\fontsize{\fsize}{#1\fsize}\selectfont}%
  \ifx\svgwidth\undefined%
    \setlength{\unitlength}{242.8344858bp}%
    \ifx\svgscale\undefined%
      \relax%
    \else%
      \setlength{\unitlength}{\unitlength * \real{\svgscale}}%
    \fi%
  \else%
    \setlength{\unitlength}{\svgwidth}%
  \fi%
  \global\let\svgwidth\undefined%
  \global\let\svgscale\undefined%
  \makeatother%
  \begin{picture}(1,0.53057059)%
    \lineheight{1}%
    \setlength\tabcolsep{0pt}%
    \put(0,0){\includegraphics[width=\unitlength,page=1]{neighborhood.pdf}}%
    \put(0.05516459,0.27812551){\color[rgb]{0,0,0}\makebox(0,0)[lt]{\lineheight{1.25}\smash{\begin{tabular}[t]{l}RL\end{tabular}}}}%
    \put(0.24538837,0.27812551){\color[rgb]{0,0,0}\makebox(0,0)[lt]{\lineheight{1.25}\smash{\begin{tabular}[t]{l}L\end{tabular}}}}%
    \put(0.37876493,0.27812551){\color[rgb]{0,0,0}\makebox(0,0)[lt]{\lineheight{1.25}\smash{\begin{tabular}[t]{l}FL\end{tabular}}}}%
    \put(0.41285413,0.11237005){\color[rgb]{0,0,0}\makebox(0,0)[lt]{\lineheight{1.25}\smash{\begin{tabular}[t]{l}FR\end{tabular}}}}%
    \put(0.23258858,0.11237005){\color[rgb]{0,0,0}\makebox(0,0)[lt]{\lineheight{1.25}\smash{\begin{tabular}[t]{l}R\end{tabular}}}}%
    \put(0.26339965,0.19534175){\color[rgb]{0,0,0}\makebox(0,0)[lt]{\lineheight{1.25}\smash{\begin{tabular}[t]{l}T\end{tabular}}}}%
    \put(0.41205833,0.19534175){\color[rgb]{0,0,0}\makebox(0,0)[lt]{\lineheight{1.25}\smash{\begin{tabular}[t]{l}F\end{tabular}}}}%
    \put(0.06311472,0.11237005){\color[rgb]{0,0,0}\makebox(0,0)[lt]{\lineheight{1.25}\smash{\begin{tabular}[t]{l}RR\end{tabular}}}}%
    \put(0,0){\includegraphics[width=\unitlength,page=2]{neighborhood.pdf}}%
    \put(0.89225926,0.18601532){\color[rgb]{0,0,0}\makebox(0,0)[lt]{\lineheight{1.25}\smash{\begin{tabular}[t]{l}Time\end{tabular}}}}%
    \put(0.74500913,0.05453896){\color[rgb]{0,0,0}\makebox(0,0)[t]{\lineheight{1.25}\smash{\begin{tabular}[t]{c}Input Channels\\$x,y,v,a$\end{tabular}}}}%
    \put(0.99889233,0.35917){\color[rgb]{0,0,0}\rotatebox{90}{\makebox(0,0)[lt]{\lineheight{1.25}\smash{\begin{tabular}[t]{l}Vehicles\end{tabular}}}}}%
    \put(0,0){\includegraphics[width=\unitlength,page=3]{neighborhood.pdf}}%
  \end{picture}%
\endgroup%

%% file: pics/architecture.pdf_tex
\begingroup%
  \makeatletter%
  \providecommand\color[2][]{%
    \errmessage{(Inkscape) Color is used for the text in Inkscape, but the package 'color.sty' is not loaded}%
    \renewcommand\color[2][]{}%
  }%
  \providecommand\transparent[1]{%
    \errmessage{(Inkscape) Transparency is used (non-zero) for the text in Inkscape, but the package 'transparent.sty' is not loaded}%
    \renewcommand\transparent[1]{}%
  }%
  \providecommand\rotatebox[2]{#2}%
  \newcommand*\fsize{\dimexpr\f@size pt\relax}%
  \newcommand*\lineheight[1]{\fontsize{\fsize}{#1\fsize}\selectfont}%
  \ifx\svgwidth\undefined%
    \setlength{\unitlength}{496.28141455bp}%
    \ifx\svgscale\undefined%
      \relax%
    \else%
      \setlength{\unitlength}{\unitlength * \real{\svgscale}}%
    \fi%
  \else%
    \setlength{\unitlength}{\svgwidth}%
  \fi%
  \global\let\svgwidth\undefined%
  \global\let\svgscale\undefined%
  \makeatother%
  \begin{picture}(1,0.24913279)%
    \lineheight{1}%
    \setlength\tabcolsep{0pt}%
    \put(0,0){\includegraphics[width=\unitlength,page=1]{architecture.pdf}}%
    \put(0.26384039,0.16468844){\color[rgb]{0,0,0}\makebox(0,0)[lt]{\lineheight{1.25}\smash{\begin{tabular}[t]{l}$24{\times}4{\times}12$\end{tabular}}}}%
    \put(0.40228759,0.17372608){\color[rgb]{0,0,0}\makebox(0,0)[lt]{\lineheight{1.25}\smash{\begin{tabular}[t]{l}$40{\times}2{\times}8$\end{tabular}}}}%
    \put(0.5698241,0.18080151){\color[rgb]{0,0,0}\makebox(0,0)[lt]{\lineheight{1.25}\smash{\begin{tabular}[t]{l}$56{\times}1{\times}4$\end{tabular}}}}%
    \put(0.70101267,0.18354527){\color[rgb]{0,0,0}\makebox(0,0)[lt]{\lineheight{1.25}\smash{\begin{tabular}[t]{l}$24{\times}1{\times}4$\end{tabular}}}}%
    \put(0,0){\includegraphics[width=\unitlength,page=2]{architecture.pdf}}%
    \put(0.75457639,0.14739747){\color[rgb]{0,0,0}\makebox(0,0)[lt]{\lineheight{1.25}\smash{\begin{tabular}[t]{l}$96{\times}1$\end{tabular}}}}%
    \put(0.74154188,0.00295754){\color[rgb]{0,0,0}\makebox(0,0)[lt]{\lineheight{1.25}\smash{\begin{tabular}[t]{l}$101{\times}1$\end{tabular}}}}%
    \put(0.87022122,0.17010497){\color[rgb]{0,0,0}\makebox(0,0)[lt]{\lineheight{1.25}\smash{\begin{tabular}[t]{l}$5{\times}1$\end{tabular}}}}%
    \put(0.11440638,0.20973038){\color[rgb]{0,0,0}\makebox(0,0)[lt]{\lineheight{1.25}\smash{\begin{tabular}[t]{l}$4{\times}8{\times}30$\end{tabular}}}}%
    \put(0.26384039,0.01969687){\color[rgb]{0,0,0}\makebox(0,0)[lt]{\lineheight{1.25}\smash{\begin{tabular}[t]{l}$24{\times}4{\times}12$\end{tabular}}}}%
    \put(0.40283198,0.02873485){\color[rgb]{0,0,0}\makebox(0,0)[lt]{\lineheight{1.25}\smash{\begin{tabular}[t]{l}$40{\times}2{\times}8$\end{tabular}}}}%
    \put(0.57036847,0.03883277){\color[rgb]{0,0,0}\makebox(0,0)[lt]{\lineheight{1.25}\smash{\begin{tabular}[t]{l}$56{\times}1\times4$\end{tabular}}}}%
    \put(0.70101267,0.04141961){\color[rgb]{0,0,0}\makebox(0,0)[lt]{\lineheight{1.25}\smash{\begin{tabular}[t]{l}$24{\times}1{\times}4$\end{tabular}}}}%
    \put(0.87022122,0.02441534){\color[rgb]{0,0,0}\makebox(0,0)[lt]{\lineheight{1.25}\smash{\begin{tabular}[t]{l}$5{\times}2$\end{tabular}}}}%
    \put(0,0){\includegraphics[width=\unitlength,page=3]{architecture.pdf}}%
    \put(-0.00014168,0.08465262){\color[rgb]{0,0,0}\makebox(0,0)[lt]{\lineheight{1.25}\smash{\begin{tabular}[t]{l}Vehicles\end{tabular}}}}%
    \put(0.0925969,0.03365848){\color[rgb]{0,0,0}\makebox(0,0)[t]{\lineheight{1.25}\smash{\begin{tabular}[t]{c}Input Channels\\$x,y,v,a$\end{tabular}}}}%
    \put(0.14004061,0.08588459){\color[rgb]{0,0,0}\makebox(0,0)[lt]{\lineheight{1.25}\smash{\begin{tabular}[t]{l}Time\end{tabular}}}}%
    \put(0,0){\includegraphics[width=\unitlength,page=4]{architecture.pdf}}%
    \put(0.29161836,0.12915324){\color[rgb]{0,0,0}\makebox(0,0)[lt]{\lineheight{1.25}\smash{\begin{tabular}[t]{l}2D-Conv + Leaky ReLU\end{tabular}}}}%
    \put(0,0){\includegraphics[width=\unitlength,page=5]{architecture.pdf}}%
    \put(0.53593125,0.12869415){\color[rgb]{0,0,0}\makebox(0,0)[lt]{\lineheight{1.25}\smash{\begin{tabular}[t]{l}Dense + Leaky ReLU\end{tabular}}}}%
    \put(0.90975068,0.21375784){\color[rgb]{0,0,0}\makebox(0,0)[lt]{\lineheight{1.25}\smash{\begin{tabular}[t]{l}Maneuver\end{tabular}}}}%
    \put(0.9109107,0.07080182){\color[rgb]{0,0,0}\makebox(0,0)[lt]{\lineheight{1.25}\smash{\begin{tabular}[t]{l}Trajectory\\\end{tabular}}}}%
    \put(0,0){\includegraphics[width=\unitlength,page=6]{architecture.pdf}}%
    \put(0.80455216,0.16618671){\color[rgb]{0,0,0}\makebox(0,0)[lt]{\lineheight{1.25}\smash{\begin{tabular}[t]{l}$40{\times}1$\end{tabular}}}}%
    \put(0.80455216,0.02310296){\color[rgb]{0,0,0}\makebox(0,0)[lt]{\lineheight{1.25}\smash{\begin{tabular}[t]{l}$40{\times}1$\end{tabular}}}}%
    \put(0,0){\includegraphics[width=\unitlength,page=7]{architecture.pdf}}%
    \put(0.9103971,0.0512721){\color[rgb]{0,0,0}\makebox(0,0)[lt]{\lineheight{1.25}\smash{\begin{tabular}[t]{l}Offsets\\\end{tabular}}}}%
    \put(0.90968871,0.19315046){\color[rgb]{0,0,0}\makebox(0,0)[lt]{\lineheight{1.25}\smash{\begin{tabular}[t]{l}Intentions\\\end{tabular}}}}%
  \end{picture}%
\endgroup%

%% file: pics/qualitative_experiment_0.pdf_tex
\begingroup%
  \makeatletter%
  \providecommand\color[2][]{%
    \errmessage{(Inkscape) Color is used for the text in Inkscape, but the package 'color.sty' is not loaded}%
    \renewcommand\color[2][]{}%
  }%
  \providecommand\transparent[1]{%
    \errmessage{(Inkscape) Transparency is used (non-zero) for the text in Inkscape, but the package 'transparent.sty' is not loaded}%
    \renewcommand\transparent[1]{}%
  }%
  \providecommand\rotatebox[2]{#2}%
  \newcommand*\fsize{\dimexpr\f@size pt\relax}%
  \newcommand*\lineheight[1]{\fontsize{\fsize}{#1\fsize}\selectfont}%
  \ifx\svgwidth\undefined%
    \setlength{\unitlength}{481.17993164bp}%
    \ifx\svgscale\undefined%
      \relax%
    \else%
      \setlength{\unitlength}{\unitlength * \real{\svgscale}}%
    \fi%
  \else%
    \setlength{\unitlength}{\svgwidth}%
  \fi%
  \global\let\svgwidth\undefined%
  \global\let\svgscale\undefined%
  \makeatother%
  \begin{picture}(1,0.10218928)%
    \lineheight{1}%
    \setlength\tabcolsep{0pt}%
    \put(0,0){\includegraphics[width=\unitlength,page=1]{qualitative_experiment_0.pdf}}%
    \put(0.02573632,0.08391584){\color[rgb]{0,0,0}\makebox(0,0)[lt]{\lineheight{1.25}\smash{\begin{tabular}[t]{l}L\end{tabular}}}}%
    \put(0.14677364,0.02903438){\color[rgb]{0,0,0}\makebox(0,0)[lt]{\lineheight{1.25}\smash{\begin{tabular}[t]{l}F\end{tabular}}}}%
    \put(0.19792482,0.08391584){\color[rgb]{0,0,0}\makebox(0,0)[lt]{\lineheight{1.25}\smash{\begin{tabular}[t]{l}FL\end{tabular}}}}%
    \put(0.29404451,0.0027009){\color[rgb]{0,0,0}\makebox(0,0)[lt]{\lineheight{1.25}\smash{\begin{tabular}[t]{l}FR\end{tabular}}}}%
    \put(0.00385847,0.03544216){\color[rgb]{0,0,0}\makebox(0,0)[lt]{\lineheight{1.25}\smash{\begin{tabular}[t]{l}T\end{tabular}}}}%
    \put(0,0){\includegraphics[width=\unitlength,page=2]{qualitative_experiment_0.pdf}}%
    \put(0.11834126,0.04135812){\color[rgb]{0,0,0}\makebox(0,0)[lt]{\lineheight{1.25}\smash{\begin{tabular}[t]{l}1\,s\end{tabular}}}}%
    \put(0.26433824,0.04420877){\color[rgb]{0,0,0}\makebox(0,0)[lt]{\lineheight{1.25}\smash{\begin{tabular}[t]{l}2\,s\end{tabular}}}}%
    \put(0.40830095,0.04751336){\color[rgb]{0,0,0}\makebox(0,0)[lt]{\lineheight{1.25}\smash{\begin{tabular}[t]{l}3\,s\end{tabular}}}}%
    \put(0,0){\includegraphics[width=\unitlength,page=3]{qualitative_experiment_0.pdf}}%
    \put(0.6219038,0.0422453){\color[rgb]{0,0,0}\makebox(0,0)[lt]{\lineheight{1.25}\smash{\begin{tabular}[t]{l}1\,s\end{tabular}}}}%
    \put(0.76853711,0.0422453){\color[rgb]{0,0,0}\makebox(0,0)[lt]{\lineheight{1.25}\smash{\begin{tabular}[t]{l}2\,s\end{tabular}}}}%
    \put(0.91175647,0.0422453){\color[rgb]{0,0,0}\makebox(0,0)[lt]{\lineheight{1.25}\smash{\begin{tabular}[t]{l}3\,s\end{tabular}}}}%
    \put(0.5293127,0.08473598){\color[rgb]{0,0,0}\makebox(0,0)[lt]{\lineheight{1.25}\smash{\begin{tabular}[t]{l}L\end{tabular}}}}%
    \put(0.70150118,0.08473598){\color[rgb]{0,0,0}\makebox(0,0)[lt]{\lineheight{1.25}\smash{\begin{tabular}[t]{l}FL\end{tabular}}}}%
    \put(0.79762091,0.00352104){\color[rgb]{0,0,0}\makebox(0,0)[lt]{\lineheight{1.25}\smash{\begin{tabular}[t]{l}FR\end{tabular}}}}%
    \put(0.50743486,0.03626229){\color[rgb]{0,0,0}\makebox(0,0)[lt]{\lineheight{1.25}\smash{\begin{tabular}[t]{l}T\end{tabular}}}}%
  \end{picture}%
\endgroup%

%% file: pics/qualitative_experiment_1.pdf_tex
\begingroup%
  \makeatletter%
  \providecommand\color[2][]{%
    \errmessage{(Inkscape) Color is used for the text in Inkscape, but the package 'color.sty' is not loaded}%
    \renewcommand\color[2][]{}%
  }%
  \providecommand\transparent[1]{%
    \errmessage{(Inkscape) Transparency is used (non-zero) for the text in Inkscape, but the package 'transparent.sty' is not loaded}%
    \renewcommand\transparent[1]{}%
  }%
  \providecommand\rotatebox[2]{#2}%
  \newcommand*\fsize{\dimexpr\f@size pt\relax}%
  \newcommand*\lineheight[1]{\fontsize{\fsize}{#1\fsize}\selectfont}%
  \ifx\svgwidth\undefined%
    \setlength{\unitlength}{490.04818726bp}%
    \ifx\svgscale\undefined%
      \relax%
    \else%
      \setlength{\unitlength}{\unitlength * \real{\svgscale}}%
    \fi%
  \else%
    \setlength{\unitlength}{\svgwidth}%
  \fi%
  \global\let\svgwidth\undefined%
  \global\let\svgscale\undefined%
  \makeatother%
  \begin{picture}(1,0.32383626)%
    \lineheight{1}%
    \setlength\tabcolsep{0pt}%
    \put(0,0){\includegraphics[width=\unitlength,page=1]{qualitative_experiment_1.pdf}}%
    \put(0.00150991,0.3090017){\color[rgb]{0,0,0}\makebox(0,0)[lt]{\lineheight{1.25}\smash{\begin{tabular}[t]{l}RL\end{tabular}}}}%
    \put(0.10802154,0.3090017){\color[rgb]{0,0,0}\makebox(0,0)[lt]{\lineheight{1.25}\smash{\begin{tabular}[t]{l}L\end{tabular}}}}%
    \put(0.12141586,0.22879237){\color[rgb]{0,0,0}\makebox(0,0)[lt]{\lineheight{1.25}\smash{\begin{tabular}[t]{l}R\end{tabular}}}}%
    \put(0.27894074,0.25251382){\color[rgb]{0,0,0}\makebox(0,0)[lt]{\lineheight{1.25}\smash{\begin{tabular}[t]{l}F\end{tabular}}}}%
    \put(0.29788257,0.3090017){\color[rgb]{0,0,0}\makebox(0,0)[lt]{\lineheight{1.25}\smash{\begin{tabular}[t]{l}FL\end{tabular}}}}%
    \put(0,0){\includegraphics[width=\unitlength,page=2]{qualitative_experiment_1.pdf}}%
    \put(0.93196266,0.28430733){\color[rgb]{0,0,0}\makebox(0,0)[lt]{\lineheight{1.25}\smash{\begin{tabular}[t]{l}5\,s\end{tabular}}}}%
    \put(0.09214238,0.25754231){\color[rgb]{0,0,0}\makebox(0,0)[lt]{\lineheight{1.25}\smash{\begin{tabular}[t]{l}T\end{tabular}}}}%
    \put(0,0){\includegraphics[width=\unitlength,page=3]{qualitative_experiment_1.pdf}}%
    \put(0.75580245,0.28419106){\color[rgb]{0,0,0}\makebox(0,0)[lt]{\lineheight{1.25}\smash{\begin{tabular}[t]{l}4\,s\end{tabular}}}}%
    \put(0,0){\includegraphics[width=\unitlength,page=4]{qualitative_experiment_1.pdf}}%
    \put(0.59448248,0.2844305){\color[rgb]{0,0,0}\makebox(0,0)[lt]{\lineheight{1.25}\smash{\begin{tabular}[t]{l}3\,s\end{tabular}}}}%
    \put(0,0){\includegraphics[width=\unitlength,page=5]{qualitative_experiment_1.pdf}}%
    \put(0.44228944,0.28420394){\color[rgb]{0,0,0}\makebox(0,0)[lt]{\lineheight{1.25}\smash{\begin{tabular}[t]{l}2\,s\end{tabular}}}}%
    \put(0,0){\includegraphics[width=\unitlength,page=6]{qualitative_experiment_1.pdf}}%
    \put(0.24816603,0.28411882){\color[rgb]{0,0,0}\makebox(0,0)[lt]{\lineheight{1.25}\smash{\begin{tabular}[t]{l}1\,s\end{tabular}}}}%
    \put(0.00151091,0.19693638){\color[rgb]{0,0,0}\makebox(0,0)[lt]{\lineheight{1.25}\smash{\begin{tabular}[t]{l}RL\end{tabular}}}}%
    \put(0.10802254,0.19693638){\color[rgb]{0,0,0}\makebox(0,0)[lt]{\lineheight{1.25}\smash{\begin{tabular}[t]{l}L\end{tabular}}}}%
    \put(0.12141685,0.11672705){\color[rgb]{0,0,0}\makebox(0,0)[lt]{\lineheight{1.25}\smash{\begin{tabular}[t]{l}R\end{tabular}}}}%
    \put(0.29788357,0.19693638){\color[rgb]{0,0,0}\makebox(0,0)[lt]{\lineheight{1.25}\smash{\begin{tabular}[t]{l}FL\end{tabular}}}}%
    \put(0,0){\includegraphics[width=\unitlength,page=7]{qualitative_experiment_1.pdf}}%
    \put(0.95100114,0.14289652){\color[rgb]{0,0,0}\makebox(0,0)[lt]{\lineheight{1.25}\smash{\begin{tabular}[t]{l}5\,s\end{tabular}}}}%
    \put(0.09214338,0.14547698){\color[rgb]{0,0,0}\makebox(0,0)[lt]{\lineheight{1.25}\smash{\begin{tabular}[t]{l}T\end{tabular}}}}%
    \put(0,0){\includegraphics[width=\unitlength,page=8]{qualitative_experiment_1.pdf}}%
    \put(0.76595085,0.17195079){\color[rgb]{0,0,0}\makebox(0,0)[lt]{\lineheight{1.25}\smash{\begin{tabular}[t]{l}4\,s\end{tabular}}}}%
    \put(0,0){\includegraphics[width=\unitlength,page=9]{qualitative_experiment_1.pdf}}%
    \put(0.60319514,0.1715169){\color[rgb]{0,0,0}\makebox(0,0)[lt]{\lineheight{1.25}\smash{\begin{tabular}[t]{l}3\,s\end{tabular}}}}%
    \put(0,0){\includegraphics[width=\unitlength,page=10]{qualitative_experiment_1.pdf}}%
    \put(0.44508943,0.17194512){\color[rgb]{0,0,0}\makebox(0,0)[lt]{\lineheight{1.25}\smash{\begin{tabular}[t]{l}2\,s\end{tabular}}}}%
    \put(0,0){\includegraphics[width=\unitlength,page=11]{qualitative_experiment_1.pdf}}%
    \put(0.25230382,0.1717687){\color[rgb]{0,0,0}\makebox(0,0)[lt]{\lineheight{1.25}\smash{\begin{tabular}[t]{l}1\,s\end{tabular}}}}%
    \put(0.10651458,0.08536154){\color[rgb]{0,0,0}\makebox(0,0)[lt]{\lineheight{1.25}\smash{\begin{tabular}[t]{l}L\end{tabular}}}}%
    \put(0.11990889,0.00515221){\color[rgb]{0,0,0}\makebox(0,0)[lt]{\lineheight{1.25}\smash{\begin{tabular}[t]{l}R\end{tabular}}}}%
    \put(0.27743376,0.02887365){\color[rgb]{0,0,0}\makebox(0,0)[lt]{\lineheight{1.25}\smash{\begin{tabular}[t]{l}F\end{tabular}}}}%
    \put(0.29637559,0.08536154){\color[rgb]{0,0,0}\makebox(0,0)[lt]{\lineheight{1.25}\smash{\begin{tabular}[t]{l}FL\end{tabular}}}}%
    \put(0,0){\includegraphics[width=\unitlength,page=12]{qualitative_experiment_1.pdf}}%
    \put(0.9399677,0.07200719){\color[rgb]{0,0,0}\makebox(0,0)[lt]{\lineheight{1.25}\smash{\begin{tabular}[t]{l}5\,s\end{tabular}}}}%
    \put(0.09063541,0.03390214){\color[rgb]{0,0,0}\makebox(0,0)[lt]{\lineheight{1.25}\smash{\begin{tabular}[t]{l}T\end{tabular}}}}%
    \put(0,0){\includegraphics[width=\unitlength,page=13]{qualitative_experiment_1.pdf}}%
    \put(0.76628651,0.06725252){\color[rgb]{0,0,0}\makebox(0,0)[lt]{\lineheight{1.25}\smash{\begin{tabular}[t]{l}4\,s\end{tabular}}}}%
    \put(0,0){\includegraphics[width=\unitlength,page=14]{qualitative_experiment_1.pdf}}%
    \put(0.6017503,0.06380322){\color[rgb]{0,0,0}\makebox(0,0)[lt]{\lineheight{1.25}\smash{\begin{tabular}[t]{l}3\,s\end{tabular}}}}%
    \put(0,0){\includegraphics[width=\unitlength,page=15]{qualitative_experiment_1.pdf}}%
    \put(0.44726951,0.06098651){\color[rgb]{0,0,0}\makebox(0,0)[lt]{\lineheight{1.25}\smash{\begin{tabular}[t]{l}2\,s\end{tabular}}}}%
    \put(0,0){\includegraphics[width=\unitlength,page=16]{qualitative_experiment_1.pdf}}%
    \put(0.25410647,0.06005594){\color[rgb]{0,0,0}\makebox(0,0)[lt]{\lineheight{1.25}\smash{\begin{tabular}[t]{l}1\,s\end{tabular}}}}%
    \put(0.4564175,0.22879237){\color[rgb]{0,0,0}\makebox(0,0)[lt]{\lineheight{1.25}\smash{\begin{tabular}[t]{l}FR\end{tabular}}}}%
    \put(0.45264388,0.11672705){\color[rgb]{0,0,0}\makebox(0,0)[lt]{\lineheight{1.25}\smash{\begin{tabular}[t]{l}FR\end{tabular}}}}%
    \put(0.45127707,0.00515221){\color[rgb]{0,0,0}\makebox(0,0)[lt]{\lineheight{1.25}\smash{\begin{tabular}[t]{l}FR\end{tabular}}}}%
    \put(0,0){\includegraphics[width=\unitlength,page=17]{qualitative_experiment_1.pdf}}%
  \end{picture}%
\endgroup%

%% file: mersch2021iros.bbl
\begin{thebibliography}{10}

\bibitem{alahi2016cvpr}
A.~Alahi, K.~Goel, V.~Ramanathan, A.~Robicquet, L.~Fei-Fei, and S.~Savarese.
\newblock {Social LSTM: Human Trajectory Prediction in Crowded Spaces}.
\newblock In {\em Proc.~of the IEEE Conf.~on Computer Vision and Pattern
  Recognition (CVPR)}, 2016.

\bibitem{amirian2019cvprws}
J.~Amirian, J.B. Hayet, and J.~Pettre.
\newblock {Social Ways: Learning Multi-Modal Distributions of Pedestrian
  Trajectories With GANs}.
\newblock In {\em Proc.~of the IEEE/CVF Conf. on Computer Vision and Pattern
  Recognition Workshops}, 2019.

\bibitem{bai2018arxiv}
S.~Bai, J.Z. Kolter, and V.~Koltun.
\newblock {An Empirical Evaluation of Generic Convolutional and Recurrent
  Networks for Sequence Modeling}.
\newblock {\em arXiv preprint}, abs/1803.01271, 2018.

\bibitem{bansal2019rss}
M.~Bansal, A.~Krizhevsky, and A.~Ogale.
\newblock {ChauffeurNet: Learning to Drive by Imitating the Best and
  Synthesizing the Worst}.
\newblock In {\em Proc.~of Robotics: Science and Systems (RSS)}, 2019.

\bibitem{brown2020arxiv}
K.~Brown, K.~Driggs-Campbell, and M.J. Kochenderfer.
\newblock {Modeling and Prediction of Human Driver Behavior: A Survey}.
\newblock {\em arXiv preprint}, abs/2006.08832, 2020.

\bibitem{casas2018corl}
S.~Casas, W.~Luo, and R.~Urtasun.
\newblock {IntentNet: Learning to Predict Intention from Raw Sensor Data}.
\newblock In {\em Proc.~of the Conf.~on Robot Learning (CoRL)}, 2018.

\bibitem{chung2014nipsws}
J.~Chung, C.~Gulcehre, K.~Cho, and Y.~Bengio.
\newblock {Empirical Evaluation of Gated Recurrent Neural Networks on Sequence
  Modeling}.
\newblock In {\em Proc.~of the Advances in Neural Information Processing
  Systems Workshops}, 2014.

\bibitem{cui2019icra}
H.~Cui, V.~Radosavljevic, F.C. Chou, T.H. Lin, T.~Nguyen, T.K. Huang,
  J.~Schneider, and N.~Djuric.
\newblock {Multimodal Trajectory Predictions for Autonomous Driving using Deep
  Convolutional Networks}.
\newblock In {\em Proc.~of the IEEE Intl.~Conf.~on Robotics \& Automation
  (ICRA)}, 2019.

\bibitem{dai2019acce}
S.~Dai, L.~Li, and Z.~Li.
\newblock {Modeling Vehicle Interactions via Modified LSTM Models for
  Trajectory Prediction}.
\newblock {\em IEEE Access}, 7:38287--38296, 2019.

\bibitem{deo2018cvprws}
N.~Deo and M.M. Trivedi.
\newblock {Convolutional Social Pooling for Vehicle Trajectory Prediction}.
\newblock {\em Proc.~of the IEEE/CVF Conf. on Computer Vision and Pattern
  Recognition Workshops}, 2018.

\bibitem{deo2018iv}
N.~Deo and M.M. Trivedi.
\newblock {Multi-Modal Trajectory Prediction of Surrounding Vehicles with
  Maneuver based LSTMs}.
\newblock In {\em Proc.~of the IEEE Vehicles Symposium (IV)}, 2018.

\bibitem{diehl2019iv}
F.~Diehl, T.~Brunner, M.~Truong-Le, and A.~Knoll.
\newblock {Graph Neural Networks for Modelling Traffic Participant
  Interaction}.
\newblock In {\em Proc.~of the IEEE Vehicles Symposium (IV)}, 2019.

\bibitem{ding2019icra}
W.~Ding, J.~Chen, and S.~Shen.
\newblock {Predicting Vehicle Behaviors Over An Extended Horizon Using Behavior
  Interaction Network}.
\newblock In {\em Proc.~of the IEEE Intl.~Conf.~on Robotics \& Automation
  (ICRA)}, 2019.

\bibitem{goodfellow2014nips}
I.J. Goodfellow, J.~Pouget-Abadie, M.~Mirza, B.~Xu, D.~Warde-Farley, S.~Ozair,
  A.~Courville, and Y.~Bengio.
\newblock {Generative Adversarial Networks}.
\newblock In {\em Proc.~of the Advances in Neural Information Processing
  Systems (NIPS)}, 2014.

\bibitem{gupta2018cvpr}
A.~Gupta, J.~Johnson, L.~Fei-Fei, S.~Savarese, and A.~Alahi.
\newblock {Social GAN: Socially Acceptable Trajectories with Generative
  Adversarial Networks}.
\newblock In {\em Proc.~of the IEEE Conf.~on Computer Vision and Pattern
  Recognition (CVPR)}, 2018.

\bibitem{halkias2006ngsim}
J.~Halkias and J.~Colyar.
\newblock {Next Generation SIMulation Fact Sheet}.
\newblock Technical report, Federal Highway Administration, 2006.

\bibitem{hochreiter1997neuralcomputation}
S.~Hochreiter and J.~Schmidhuber.
\newblock {Long Short-Term Memory}.
\newblock {\em Neural Computation}, 9(8):1735--1780, 1997.

\bibitem{hoermann2018icra}
S.~Hoermann, M.~Bach, and K.~Dietmayer.
\newblock {Dynamic Occupancy Grid Prediction for Urban Autonomous Driving: A
  Deep Learning Approach with Fully Automatic Labeling}.
\newblock In {\em Proc.~of the IEEE Intl.~Conf.~on Robotics \& Automation
  (ICRA)}, 2018.

\bibitem{hong2019cvpr}
J.~Hong, B.~Sapp, and J.~Philbin.
\newblock {Rules of the Road: Predicting Driving Behavior with a Convolutional
  Model of Semantic Interactions}.
\newblock In {\em Proc.~of the IEEE/CVF Conf.~on Computer Vision and Pattern
  Recognition (CVPR)}, 2019.

\bibitem{hu2018iv}
Y.~Hu, W.~Zhan, and M.~Tomizuka.
\newblock {Probabilistic Prediction of Vehicle Semantic Intention and Motion}.
\newblock In {\em Proc.~of the IEEE Vehicles Symposium (IV)}, 2018.

\bibitem{kayhan2020cvpr}
O.S. Kayhan and J.C. van Gemert.
\newblock {On Translation Invariance in CNNs: Convolutional Layers can Exploit
  Absolute Spatial Location}.
\newblock In {\em Proc.~of the IEEE/CVF Conf.~on Computer Vision and Pattern
  Recognition (CVPR)}, 2020.

\bibitem{kim2017itsc}
B.~Kim, C.~M. Kan, J.~Kim, S.~H. Lee, C.~C. Chung, and J.~W. Choi.
\newblock {Probabilistic Vehicle Trajectory Prediction over Occupancy Grid Map
  via Recurrent Neural Network}.
\newblock In {\em Proc.~of the IEEE Intl.~Conf.~on Intelligent Transportation
  Systems (ITSC)}, 2017.

\bibitem{kingma2015iclr}
D.P. Kingma and J.~Ba.
\newblock {Adam: {A} Method for Stochastic Optimization}.
\newblock In {\em Proc.~of the Int.~Conf.~on Learning Representations (ICLR)},
  2015.

\bibitem{krajewski2018itsc}
R.~Krajewski, J.~Bock, L.~Kloeker, and L.~Eckstein.
\newblock {The highD Dataset: A Drone Dataset of Naturalistic Vehicle
  Trajectories on German Highways for Validation of Highly Automated Driving
  Systems}.
\newblock In {\em Proc.~of the IEEE Intl.~Conf.~on Intelligent Transportation
  Systems (ITSC)}, 2018.

\bibitem{kuefler2017iv}
A.~Kuefler, J.~Morton, T.~A. Wheeler, and M.J. Kochenderfer.
\newblock {Imitating Driver Behavior with Generative Adversarial Networks}.
\newblock In {\em Proc.~of the IEEE Vehicles Symposium (IV)}, 2017.

\bibitem{lefevre2014robo}
S.~Lef{\`{e}}vre, D.~Vasquez, and C.~Laugier.
\newblock {A survey on motion prediction and risk assessment for intelligent
  vehicles}.
\newblock {\em Journal of Robotics and Mechanical Engineering Research},
  1:1--14, 2014.

\bibitem{li2020iros}
L.L. Li, B.~Yang, M.~Liang, W.~Zeng, and M.~Ren.
\newblock {End-to-end Contextual Perception and Prediction with Interaction
  Transformer}.
\newblock {\em Proc.~of the IEEE/RSJ Intl.~Conf.~on Intelligent Robots and
  Systems (IROS)}, 2020.

\bibitem{luo2018cvpr}
W.~Luo, B.~Yang, and R.~Urtasun.
\newblock {Fast and Furious: Real Time End-to-End 3D Detection, Tracking and
  Motion Forecasting with a Single Convolutional Net}.
\newblock In {\em Proc.~of the IEEE Conf.~on Computer Vision and Pattern
  Recognition (CVPR)}, 2018.

\bibitem{maas2013icml}
A.~Maas, A.Y. Hannun, and A.Y. Ng.
\newblock {Rectifier Nonlinearities Improve Neural Network Acoustic Models}.
\newblock In {\em Proc.~of the Int.~Conf.~on Machine Learning (ICML)}, 2013.

\bibitem{mozaffari2020arxiv}
S.~Mozaffari, O.Y. Al-Jarrah, M.~Dianati, P.~Jennings, and A.~Mouzakitis.
\newblock {Deep Learning-Based Vehicle Behavior Prediction for Autonomous
  Driving Applications: A Review}.
\newblock {\em arXiv preprint}, abs/1912.11676, 2019.

\bibitem{nikhil2018eccvws}
N.~Nikhil and B.T. Morris.
\newblock {Convolutional Neural Network for Trajectory Prediction}.
\newblock In {\em Proc.~of the Europ.~Conf.~on Computer Vision Workshops},
  2018.

\bibitem{park2018iv}
S.H. Park, B.~Kim, C.M. Kang, C.C. Chung, and J.W. Choi.
\newblock {Sequence-to-Sequence Prediction of Vehicle Trajectory via LSTM
  Encoder-Decoder Architecture}.
\newblock In {\em Proc.~of the IEEE Vehicles Symposium (IV)}, 2018.

\bibitem{sadeghian2019cvpr}
A.~Sadeghian, V.~Kosaraju, A.~Sadeghian, N.~Hirose, S.~H. Rezatofighi, and
  S.~Savarese.
\newblock {SoPhie: An Attentive GAN for Predicting Paths Compliant to Social
  and Physical Constraints}.
\newblock In {\em Proc.~of the IEEE/CVF Conf.~on Computer Vision and Pattern
  Recognition (CVPR)}, 2019.

\bibitem{schulz2018iros}
J.~Schulz, C.~Hubmann, J.~Lochner, and D.~Burschka.
\newblock {Interaction-Aware Probabilistic Behavior Prediction in Urban
  Environments}.
\newblock In {\em Proc.~of the IEEE/RSJ Intl.~Conf.~on Intelligent Robots and
  Systems (IROS)}, 2018.

\bibitem{song2020eccv}
H.~Song, W.~Ding, Y.~Chen, S.~Shen, M.~Wang, and Q.~Chen.
\newblock {PiP: Planning-informed Trajectory Prediction for Autonomous
  Driving}.
\newblock {\em Proc.~of the Europ.~Conf.~on Computer Vision (ECCV)}, 2020.

\bibitem{sun2019iv}
L.~Sun, W.~Zhan, C.Y. Chan, and M.~Tomizuka.
\newblock {Behavior Planning of Autonomous Cars with Social Perception}.
\newblock In {\em Proc.~of the IEEE Vehicles Symposium (IV)}, 2019.

\bibitem{tang2019neurips}
Y.~Tang and R.~Salakhutdinov.
\newblock {Multiple Futures Prediction}.
\newblock {\em Proc.~of the Conference on Neural Information Processing Systems
  (NeurIPS)}, 2019.

\bibitem{xin2018itsc}
L.~Xin, P.~Wang, C.Y. Chan, J.~Chen, S.E. Li, and B.~Cheng.
\newblock {Intention-aware Long Horizon Trajectory Prediction of Surrounding
  Vehicles using Dual LSTM Networks}.
\newblock In {\em Proc.~of the IEEE Intl.~Conf.~on Intelligent Transportation
  Systems (ITSC)}, 2018.

\bibitem{yu2016iclr}
F.~Yu and V.~Koltun.
\newblock {Multi-Scale Context Aggregation by Dilated Convolutions}.
\newblock In {\em Proc.~of the Int.~Conf.~on Learning Representations (ICLR)},
  2016.

\bibitem{zhao2019cvpr}
T.~Zhao, Y.~Xu, M.~Monfort, W.~Choi, C.~Baker, Y.~Zhao, Y.~Wang, and Y.N. Wu.
\newblock {Multi-Agent Tensor Fusion for Contextual Trajectory Prediction}.
\newblock In {\em Proc.~of the IEEE/CVF Conf.~on Computer Vision and Pattern
  Recognition (CVPR)}, 2019.

\bibitem{zyner2018tits}
A.~Zyner, S.~Worrall, and E.~Nebot.
\newblock {Naturalistic Driver Intention and Path Prediction Using Recurrent
  Neural Networks}.
\newblock {\em IEEE Trans.~on Intelligent Transportation Systems (ITS)},
  21(4):1584--1594, 2018.

\end{thebibliography}
